\def\year{2021}\relax
\documentclass[letterpaper]{article} 
\usepackage{aaai21}  
\usepackage{times}  
\usepackage{helvet} 
\usepackage{courier}  
\usepackage[hyphens]{url}  
\usepackage{graphicx} 
\usepackage{natbib}  
\usepackage{amsmath,amssymb} 
\usepackage{caption} 
\usepackage{color}
\usepackage{algorithm,epstopdf,algorithmic,subfigure,booktabs,ctable}
\title{BSN++: Complementary Boundary Regressor with Scale-Balanced Relation Modeling for Temporal Action Proposal Generation}
\author{
    Haisheng Su\textsuperscript{\rm 1}\thanks{Corresponding author.},
        Weihao Gan\textsuperscript{\rm 1}, 
        Wei Wu\textsuperscript{\rm 1},
        Yu Qiao\textsuperscript{\rm 2, 3},
        Junjie Yan\textsuperscript{\rm 1}
    \\
}
\affiliations{
    \textsuperscript{\rm 1} SenseTime Research \\
    \textsuperscript{\rm 2} Shenzhen Institutes of Advanced Technology, Chinese Academy of Sciences \\
    \textsuperscript{\rm 3}  Shanghai AI Laboratory, Shanghai, China \\
    \{suhaisheng, ganweihao, wuwei, yanjunjie\}@sensetime.com, 
    yu.qiao@siat.ac.cn\\
}

\begin{document}


\maketitle

\begin{abstract}
Generating human action proposals in untrimmed videos is an important yet challenging task with wide applications. Current methods often suffer from the noisy boundary locations and the inferior quality of confidence scores used for proposal retrieving. In this paper, we present BSN++, a new framework which exploits complementary boundary regressor and relation modeling for temporal proposal generation. First, we propose a novel boundary regressor based on the complementary characteristics of both starting and ending boundary classifiers. Specifically, we utilize the U-shaped architecture with nested skip connections to capture rich contexts and introduce bi-directional boundary matching mechanism to improve boundary precision. Second, to account for the proposal-proposal relations ignored in previous methods, we devise a proposal relation block to which includes two self-attention modules from the aspects of position and channel. Furthermore, we find that there inevitably exists data imbalanced problems in the positive/negative proposals and temporal durations, which harm the model performance on tail distributions. To relieve this issue, we introduce the scale-balanced re-sampling strategy. Extensive experiments are conducted on two popular benchmarks: ActivityNet-1.3 and THUMOS14, which demonstrate that BSN++ achieves the state-of-the-art performance. Not surprisingly, the proposed BSN++ ranked $1^{st}$ place in the CVPR19 - ActivityNet challenge leaderboard on temporal action localization task. 


\end{abstract}


\section{Introduction}

\begin{figure}[t]
	\centering
	\includegraphics[width=0.9\columnwidth]{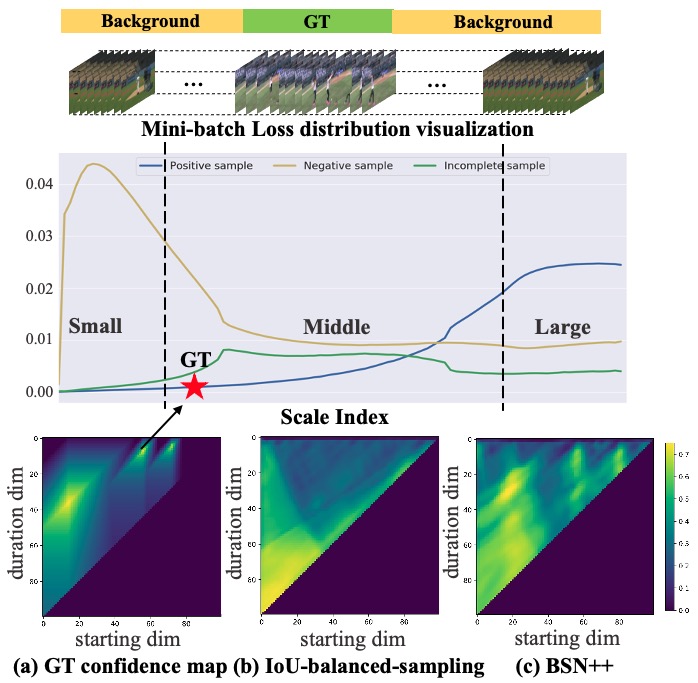} 
	\caption{(a) Given an untrimmed video containing several action instances of small scale, (b) IoU-balanced sampling is widely used to train the proposal confidence regressor, which still suffers from inferior quality owing to the imbalanced distribution of the temporal durations, resulting in the long-tailed proposal dataset. (c) BSN++ aims at generating high-quality proposal boundaries as well as reliable confidence scores with complementary boundary regressor and scale-balanced proposal relation block. }
	\label{fig:overview}
\end{figure}

Temporal action detection task has received much attention from many researchers in recent years, which requires not only categorizing the real-world untrimmed videos but also locating the temporal boundaries of action instances. Akin to object proposals for object detection in images, temporal action proposal indicates the temporal intervals containing the actions and plays an important role in temporal action detection. It has been commonly recognized that high-quality proposals usually have two crucial properties: (1) the generated proposals should cover the action instances temporally with both high recall and temporal overlapping; (2) the quality of proposals should be evaluated accurately, thus providing a overall confidence for later retrieving step.

To cater for these two conditions and achieve high quality proposals, there are two main categories in the existing proposal generation methods \cite{sst_buch_cvpr17,gao2017turn,SSAD,SCNN}. The first type adopts the \textit{top-down} fashion, where proposals are generated based on sliding windows \cite{SCNN} or uniform-distributed anchors \cite{SSAD}, then a binary classifier is employed to evaluate confidence for the proposals. 
However, the proposals generated in this way are doomed to have imprecise boundaries though with regression. Under this circumstance, the other type of methods \cite{BSN,Y.Xiong,LinBMN} attract many researchers recently which tackle this problem in a \textit{bottom-up} fashion, where the input video is evaluated in a finer-level. \cite{BSN} is a typical method in this type which proposes the Boundary-Sensitive Network (BSN) to generate proposals with flexible durations and reliable confidence scores. Though BSN achieves convincing performance, it still suffers from three main drawbacks: (1) BSN only employs the local details around the boundaries to predict boundaries, without taking advantage of the rich temporal contexts through the whole video sequence; (2) BSN fails to consider the proposal-proposal relations for confidence evaluation; (3) the imbalance data distribution between positive/negative proposals and temporal durations is also neglected.

To relieve these issues, we propose BSN++, for temporal proposal generation. \textbf{(i)} To exploit the rich contexts for boundary prediction, we adopt the U-shaped architecture with nested skip connections. Meanwhile, the two optimized boundary classifiers share the same goals especially in detecting the sudden change from background to actions or learning the discriminativeness from actions to background, thus are complementary with each other. Under this circumstance, we propose the complementary boundary regressor, where the starting classifier can also be used to predict the ending locations when the input videos are processed in a reversed direction, and vice versa. In this way, we can achieve high precision without adding extra parameters. \textbf{(ii)} In order to predict the confidence scores of densely-distributed proposals, we design a proposal relation block aiming at leveraging both channel-wise and position-wise global dependencies for proposal-proposal relation modeling. \textbf{(iii)} To relieve the imbalance scale-distribution among the sampling positives as well as the negatives (see Fig. \ref{fig:overview}), we implement a two-stage re-sampling scheme consisting of the IoU-balanced (positive-negative) sampling and the scale-balanced re-sampling. The boundary map and the confidence map are generated simultaneously and jointly trained in a unified framework. In summary, the main contributions of our work are listed below in three-folds:

\begin{itemize}
	\item We revisit the boundary prediction problem and propose a complementary boundary generator to exploit both ``\textit{local and global}", ``\textit{past and future}" contexts for accurate temporal boundary prediction.
	
	\item We propose a proposal relation block for proposal confidence evaluation, where two self-attention modules are adopted to model the proposal relations from two complementary aspects. Besides, we devise a two-stage re-sampling scheme for equivalent balancing. 
	
	\item Thorough experiments are conducted to reveal the effectiveness of our method. Further combining with the existing action classifiers, our method can achieve the state-of-the-art temporal action detection performance.
	
\end{itemize}


\section{Related Work}
\subsection{Action Recognition}
Action recognition is an essential branch which has been extensively explored in recent years. Earlier methods such as improved Dense Trajectory (iDT) \cite{DT,iDT} mainly adopt the hand-crafted features including HOG, MBH and HOF. Current deep learning based methods \cite{C.Feichtenhofer,K.Simonyan,D.Tran,TSN,su2020collaborative} typically contain two main categories: the two-stream networks \cite{C.Feichtenhofer,K.Simonyan} capture the appearance and motion information from RGB image and stacked optical flow respectively; 3D networks \cite{D.Tran,p3d} exploit 3D convolutions to capture the spatial and temporal information directly from the raw videos. Action recognition networks are usually adopted to extract visual feature sequence from untrimmed videos for the temporal action proposals and detection task.

\subsection{Imbalanced Distribution Training} 
Imbalanced data distribution naturally exists in many large-scale datasets \cite{OpenImage,cityscapes,Feng_2018_CVPR_Workshops}. Current literature can be mainly divided into three categories: (1) re-sampling, includes oversampling the minority classes \cite{Andrew,ji2020context} or downsampling the majority classes \cite{Gary,hu2020class}; (2) re-weighting, namely cost sensitive learning \cite{Kate,Cui}, which aims to dynamically adjust the weight of samples or different classes during training process. (3) In object detection task, the imbalanced issue is more serious between background and foreground for one-stage detector. Some methods such as Focal loss \cite{TsungYi} and online hard negative mining \cite{Abhinav} are designed for two-stage detector. In this paper, we implement the scale-balanced re-sampling upon the IoU-balanced sampling for proposal confidence evaluation, motivated by the mini-batch imbalanced loss distribution against proposal durations.

\subsection{Temporal Action Detection and Proposals}

Akin to object detection in images, temporal action detection also can be divided
into proposal and classification stages. Current methods train these two stages separately \cite{G.Singh} or jointly \cite{sstad,SSAD}. Top-down methods \cite{SSAD} are mainly based on sliding windows or pre-defined anchors, while bottom-up methods \cite{Y.Xiong,BSN,LinBMN} first evaluate the actionness or boundary probabilities of each temporal location in a finer level. 
However, proposals generated in a local fashion of \cite{Y.Xiong} cannot be further retrieved without confidence scores evaluated from a global view. And probabilities sequence generated in \cite{BSN,LinBMN,LiuMulti} is sensitive to noises, causing many false alarms. Besides, proposal-proposal relations fail to be considered for confidence evaluation. Meanwhile, the imbalanced distribution among the proposals remains to be settled. To address these issues, we propose BSN++, which is unique to previous works in three main aspects: (1) we revisit the boundary prediction task and propose to exploit rich contexts together with bi-directional matching strategy for accurate boundary prediction; (2) we devise a proposal relation block for proposal-proposal relations modeling; (3) two-stage re-sampling scheme is designed for equivalent balancing.

\begin{figure}[t]
	\centering
	\includegraphics[width=0.36\textwidth]{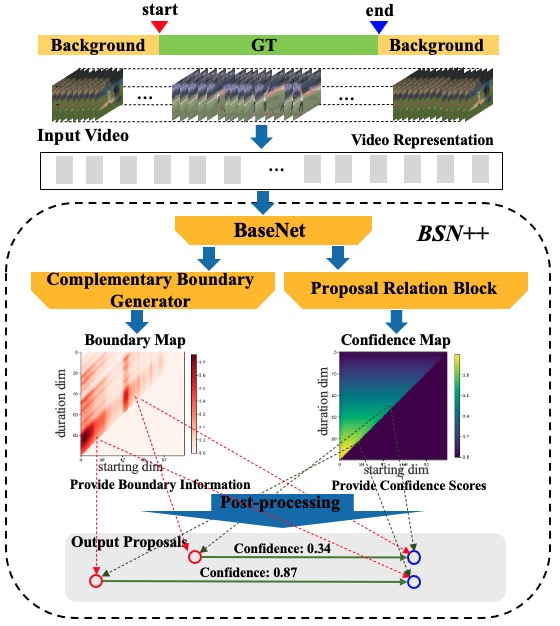} 
	\caption{The framework of BSN++. Given an untrimmed video, two-stream network is adopted to extract visual features. Then BSN++ can densely evaluate all proposals by producing the boundary map with a complementary boundary generator and the confidence map with a proposal relation block simultaneously.}
	\label{fig:framework}
\end{figure}


\section{Our Approach}


\subsection{Problem Definition}

Denote an untrimmed video sequence as $ \mathbf{U} = \{\mathbf{u}_{t}\}_{t=1}^{l_{v}}, $ where $ \mathbf{u}_{t} $ indicates the $ t $-th frame in the video of length $ l_{v} $. A set of action instances $ \mathbf{\Psi}_{g} = \{\mathbf{\varphi}_{n} = (t_{n}^{s}, t_{n}^{e})\}_{n=1}^{N_{g}} $ are temporally annotated in the video $  \mathbf{S}_{v} $, where $ N_{g} $ is the number of ground truth action instances, and $ t_{n}^{s}, t_{n}^{e} $ are the starting time and ending time of the action instance $ \varphi_{n} $ respectively. During training phase, the $ \mathbf{\Psi}_{g} $ is provided. While in the testing phase, the predicted proposal set $ \mathbf{\Psi}_{p} $ should cover the $ \mathbf{\Psi}_{g} $ with high recall and high temporal overlapping.

\subsection{Video Feature Encoding}

Before applying our algorithm, we adopt the two-stream network \cite{K.Simonyan} in advance to encode the visual features from raw video as many previous works \cite{BSN,Gao2018CTAP,Su2018Cascaded,Su2020MGFN}. This kind of architecture has been widely used in many video analysis tasks\cite{SSAD,SSN,CBR}. Concretely, given an untrimmed video $ \mathbf{S}_{v} $ which contains $ l_{v} $ frames, we process the input video in a regular interval $ \sigma $ for reducing the computational cost. We concatenate the output of the last FC-layer in the two-stream network to form the feature sequence $ \mathbf{F}=\{\mathbf{f}_{i}\}_{i=1}^{l_{s}}$, where $ l_{s} = l_{v}/\sigma $. Final, the feature sequence $ \mathbf{F}$ is used as the input of our BSN++.

\begin{figure}[t]
	\centering
	\includegraphics[width=0.48\textwidth]{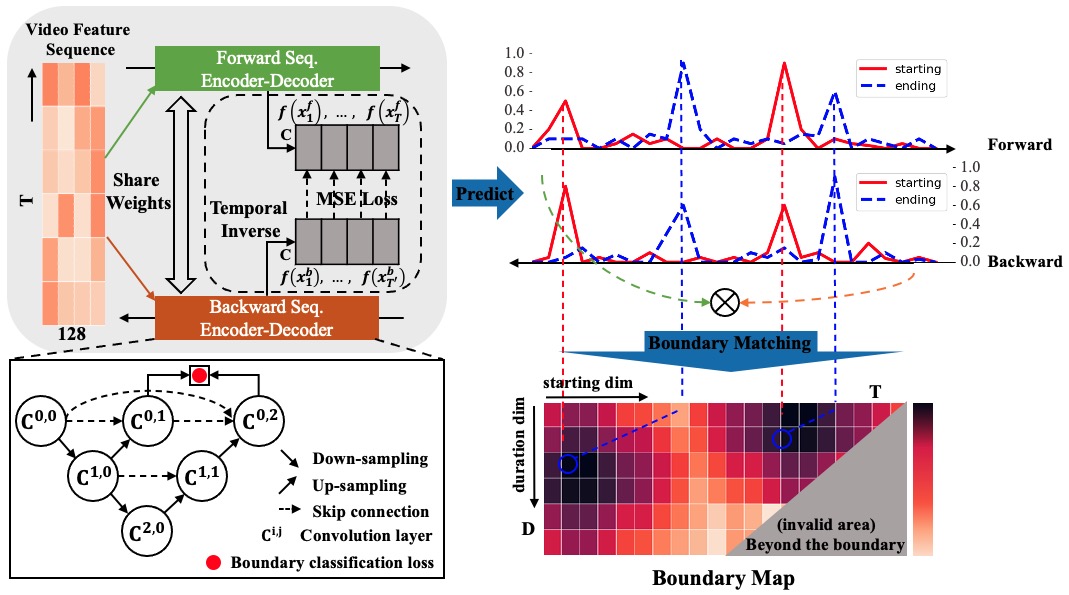} 
	\caption{Illustration of the complementary boundary generator. U-shaped encoder-decoder with dense skip connections are utilized for accurate boundary prediction. Consistent regularization is performed on the intermediate features during the training process. In inference stage, starting/ending classifiers are also utilized to predict the ending/starting locations in a backward order. The two siamese backbones share the weights. Finally the boundary map is constructed through matching boundary locations into pairs based on the two-passes boundary probabilities sequence.}
	\label{fig:TEM}
\end{figure}

\subsection{Proposed Network Architecture: BSN++}

In contrast to the previous BSN \cite{BSN}, which consists of multiple stages, BSN++ is designed to generate the proposal map directly in a unified network. To obtain the proposal map, BSN++ first generates the boundary map which represents the boundary information and confidence map which represents the confidence scores of densely distributed proposals. As shown in Fig. \ref{fig:framework}, BSN++ model mainly contains three main modules:  \textit{Base Module} handles the input video features to perform temporal information modeling, then the output features are shared by the two following modules. \textit{Complementary Boundary Generator} processes the input video features to evaluate the boundary probabilities sequence, using a nested U-shaped encoder-decoder; \textit{Proposal Relation Block} aims to model the proposal-proposal relations with two self-attention modules responsible for two complementary dependencies.

\noindent
\textbf{Base Module.} The goal of this module is to handle the extracted features for temporal relationship modeling, which serves as the base module of the following two branches. It mainly includes two 1D convolutional layers, with 256 filters, kernel size 3 and stride 1, followed by a ReLU activation layer. Since the length of videos is uncertain, we truncate the video sequence into a series of sliding windows. The detailed of data construction is illustrated in Section 4.1.

\noindent
\textbf{Complementary Boundary Generator.}  Inspired by the success of U-Net \cite{unet,Unet++} used in image segmentation, we design our boundary generator as Encoder-Decoder networks because this kind of architecture is able to capture both high-level \textit{global} context and low-level \textit{local} details at the same time. As shown in Fig. \ref{fig:TEM}, each circle represents a 1D convolutional layer with 512 filters and kernel size 3, stride 1, together with a batch normalization layer and a ReLU layer except the prediction layer. To reduce over-fitting, we just add two down-sampling layers to expand the receptive fields and the same number of up-sampling layers are followed to recover the original temporal resolutions. Besides, deep supervision (shown {\color{red}red}) is also performed for fast convergent speed and nested skip connections are employed for bridging the semantic gap between feature maps of the encoder and decoder prior to fusion.


We observe that the starting classifier learns to detect the sudden change from background to actions and vice versa. Hence, the starting classifier can be regarded as a pseudo ending classifier when processes the input video in a reversed direction, thus the bi-directional prediction results are complementary. With this observation, bi-directional encoder-decoder networks are optimized in parallel, and the consistent constraint is performed upon the intermediate features (i.e. $f(x^{f}_{i})$ and  $f(x^{b}_{i})$) on both sides before the prediction layer as shown in Fig. \ref{fig:TEM}. During the inference stage, the aforementioned encoder-decoder network is adopted to predict the 
the starting heatmap $ \overrightarrow{\mathbf{H}}^{s}=\{\overrightarrow{\mathbf{h}}_{i}^{s}\}_{i=1}^{l_{s}} $ and ending heatmap $ \overrightarrow{\mathbf{H}}^{e}=\{\overrightarrow{\mathbf{h}}_{i}^{e}\}_{i=1}^{l_{s}} $ respectively,  where $ \mathbf{h}_{i}^{s}$ and $\mathbf{h}_{i}^{e}$ indicate the starting and ending probabilities of the $ i $-th snippet respectively. Meanwhile, we feed the input feature sequence in a reversed order to the identical backbone. Similarly, we can obtain the starting heatmap $ \overleftarrow{\mathbf{H}}^{s}$ and ending heatmap $ \overleftarrow{\mathbf{H}}^{e}$. 

After the two-passes, in order to select the boundaries of high scores, we fuse the two pairs of heatmaps to yield the final heatmaps:
\begin{equation}
	\mathbf{H}^{s} = \{\sqrt{\overrightarrow{\mathbf{h}_{i}^{s}} \times \overleftarrow{\mathbf{h}}_{i}^{s}}\}_{i=1}^{l_{s}},
	\mathbf{H}^{e} = \{\sqrt{\overrightarrow{\mathbf{h}}_{i}^{e} \times \overleftarrow{\mathbf{h}}_{i}^{e}}\}_{i=1}^{l_{s}},
\end{equation}


With these two boundary points heatmaps, we can further construct the boundary map $\mathbf{M}^{b} \in R^{1\times D\times T} $ which can represent the boundary information of all densely distributed proposals, where $ T $ and $ D $ are the length of the feature sequence and maximum duration of proposals separately:
\begin{equation}
	\mathbf{M}_{j,i}^{b} = \{\{\mathbf{h}_{i}^{s} \times\mathbf{h}_{i+j}^{e}\}_{i=1}^{T}\}_{j=1}^{D},   i + j < T,
\end{equation}

\noindent
\textbf{Proposal Relation Block.} The goal of this block is to evaluate the confidence scores of dense  proposals. Before performing proposal-proposal relations, we follow the previous work BMN \cite{LinBMN} to generate the proposal feature maps as $\mathbf{F}^{p} \in R^{D\times T\times 128\times N } $. N is set to 32. Then the proposal feature maps are fed to a 3D convolutional layer with kernel size $ 1\times 1\times 32 $ and 512 filters, followed by a ReLU activation layer. Thus the reduced proposal features maps are $\widehat{\mathbf{F}^{p}} \in R^{D\times T\times 512} $. The proposal relation block consists of two self-attention modules as follows.

\begin{figure}[t]
	\begin{center} 
		\includegraphics[width=0.95\linewidth]{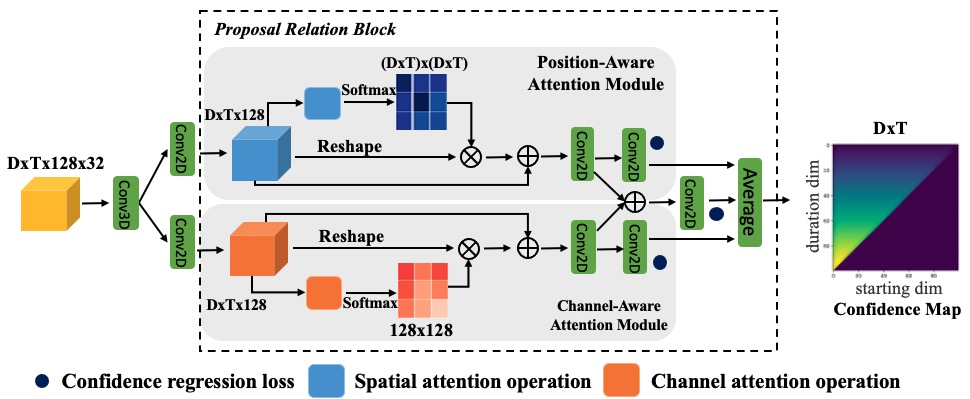}
	\end{center}
	\caption{Illustration of the proposal relation block. After generating the proposal feature maps, two complementary branches are followed to model the proposal relation separately. In the upper branch, the position-aware attention module aims to leverage \textit{global} dependencies. While in the lower branch, the channel-aware attention module aims to attend to the discriminative features by channel matrix calculation. Finally, we aggregate the outputs from the three branches for \textit{pixel-level} confidence prediction.}
	\label{fig:PEM}
\end{figure}   

\textbf{Position-aware attention module.} As illustrated in Fig. \ref{fig:PEM}, given the proposal features $\widehat{\mathbf{F}^{p}}$, we adopt the similar self-attention mechanism as \cite{nnn}, where the proposal feature maps are fed into a convolutional layer separately to generate 
two new feature maps $ A $ and $ B $ for spatial matrix multiplication with reshape and transpose operations. And then a Softmax layer is applied to calculate the position-aware attention $ P^{A}\in R^{L\times L} $, where $ L=D\times T $:
\begin{equation}
P^{A}_{j,i} = \dfrac{\exp(A_{i}\cdot B_{j})}{\sum_{i=1}^{L}\exp(A_{i}\cdot B_{j})} ,
\end{equation}
where $ P^{A}_{j,i} $ indicates the attention of the $ i^{th} $ position on the $ j^{th} $ position.  Finally, the attended features are further weighted summed with the proposal features and fed to the convolutional layers for confidence prediction.

\textbf{Channel-aware attention module.} In contrast to the position-aware attention module, this module directly performs channel-wise matrix multiplication in order to exploit the inter-dependencies among different channels, which can help enhance the proposal feature representations for confidence prediction. The process of attention calculation is the same as the former module except for the attended dimension. Similarly, the attended features after weighted summed with the proposal features are further captured by a 2D convolutional layer to generate the confidence map $ \mathbf{M}^{c} \in R^{D\times T} $. We also aggregate the outputs of the two attention modules for proposal confidence prediction, and finally we fuse the predicted confidence maps from the three branches for a better performance.


\subsection{Re-sampling}

Imbalanced data distribution can affect the model training especially in the long-tailed dataset. In this paper, we revisit the positive/negative samples distribution for improving the quality of proposal confidence prediction and design a proposal-level re-sampling method to improve the performance of training on the long-tailed dataset. Our re-sampling scheme consists of two stages aiming at not only balancing the positives and negatives proposals, but also balancing the temporal duration of the proposals.
\subsubsection{IoU-balanced sampling} As shown in Fig. \ref{fig:overview}, we can see from the mini-batch loss distribution that the number of positives and negatives differs greatly which dooms to bias the training model without effective measures. Previous works usually design a positive-negative sampler (i.e. IoU-balanced sampler) to balance the data distribution for each mini-batch, thus ensuring the ratio of positive and negative samples is nearly 1:1. However, we can also conclude from the Fig. \ref{fig:overview} that the scale of positives or negatives fails to conform the uniform distribution. Under this circumstance, we should consider how to balance the scales of proposals.
\subsubsection{Scale-balanced re-sampling} To relieve the issue among long-tailed scales, we propose a second-stage positive/negative re-sampling method, which is upon the principle of IoU-balanced sampling. Specifically, define $ P_{i} $ as the number of positive proposals with the scale $ s_{i} $, then $ r_{i} $ is the positive ratio of $ s_{i} $:
\begin{equation}
\begin{split}
   & r_{i} = \dfrac{P_{i}}{\sum_{j=1}^{N_{s}} P_{j}} , \\
   & r^{'}_{i} = \left\{  
   \begin{aligned}
             & \lambda \ast exp^{(\dfrac{r_{i}}{\lambda} - 1)} \quad & (0 < r_{i} \leq \lambda ) ,  \\  
             &  r_{i} \quad &(\lambda < r_{i} \leq 1 ) , 
   \end{aligned}  
\right. 
\end{split}
\end{equation}
where $ N_{s} $ is the number of pre-defined normalized scale regions (i.e. $[0 - 0.3, 0.3 - 0.7, 0.7 - 1.0]$). Then we design a positive ratio sampling function, the resulting ratio $  r^{'}_{i}$ is bigger than $  r_{i}$ for proposal scale with a frequency lower than $ \lambda$, where $ \lambda$ is a hyper-parameter which we set to 0.15 empirically. Hence, we use the re-normalized $  r^{'}_{i}$ as the sampling probability of the specific proposal scale region $ s_{i} $ to construct the mini-batch data. As for the negative proposals, the same process is performed.

\section{Training and Inference of BSN++}


\subsection{Training}
\textbf{Overall Objective Function.} As described above, BSN++ consists of three main sub-modules. The multi-task objective function is defined as:
\begin{equation}
	L_{BSN++} = L_{CBG} + \beta\cdot L_{PRB} + \gamma\cdot L_{2}(\Theta),
\end{equation}
where $ L_{CBG} $ and $ L_{PRB} $ are the objective functions of the complementary boundary generator and the proposal relation block respectively, while $ L_{2}(\Theta)$ is a regularization term. $\beta $ and $ \gamma $ are set to 10 and 0.0001 separately to trade off the training process of two modules and reduce over-fitting. 

\noindent
\textbf{Training Data Construction.} Given the extracted feature $\mathbf{ F }$ with length $ l_{s} $, we truncate $\mathbf{ F }$ into sliding windows of length $ l_{w} $ with 75\% temporal overlapping. Then we construct the training dataset as $ \Phi = \{\mathbf{ F }^{w}_{n}\}_{n=1}^{N_{w}} $, where $ N_{w} $ is the number of retained windows containing at least one ground-truth. 	

\noindent
\textbf{Label Assignment.} For the Complementary Boundary Generator (CBG), in order to predict the boundary probabilities sequence, we need to generate the corresponding label sequence $ \mathbf{G}_{s}^{w} $ and $ \mathbf{G}_{e}^{w} $ as in \cite{BSN}. Specifically, for each action instance $ \varphi_{g} $ in the annotation set $ \mathbf{\Psi}_{g}^{w} $, we denote it's starting and ending regions as $ [t_{g}^{s}-d_{\varphi}/10, t_{g}^{s}+d_{\varphi}/10] $ and $ [t_{g}^{e}-d_{\varphi}/10, t_{g}^{e}+d_{\varphi}/10] $ respectively, where $ d_{\varphi}= t_{g}^{e} - t_{g}^{s} $ is the duration of $ \varphi_{g} $. Then for each temporal location, if it lies in the starting or ending regions of any action instances, the corresponding label $ \mathbf{g}^{s} $ or $ \mathbf{g}^{e} $ will be set to 1. Hence the label sequence of starting and ending used in CBG are $ \mathbf{G}_{s}^{w}=\{\mathbf{g}_{i}^{s}\}_{i=1}^{l_{w}} $, $ \mathbf{G}_{e}^{w}=\{\mathbf{g}_{i}^{e}\}_{i=1}^{l_{w}} $ respectively.

For the Proposal Relation Block (PRB), we predict the confidence map $\mathbf{M}^{c} \in R^{D\times l_{w}} $ of all densely distributed proposals, where the point $ g^{c}_{j,i} $ in the label confidence map $\mathbf{M}_{g}^{c} = \{\{ g^{c}_{j,i} \}_{i=1}^{l_{w}}\}_{j=1}^{D} $ represents the maximum $ IoU $ (Intersection-over-Union) values  of proposal $ \varphi_{j,i} = [t_{s} = i, t_{e} = i+j] $ with all $ \varphi_{g} $ in $ \mathbf{\Psi}_{g}^{w} $.

\noindent
\textbf{Objective of CBG.} We follow \cite{BSN} to adopt the weighted binary logistic regression loss $ L_{bl} $ as the objective between the output probability and the corresponding label sequence. The objective is:
\begin{equation}
	L_{CBG} = \underbrace{\overrightarrow{L_{bl}^{s}} + \overrightarrow{L_{bl}^{e}}}_{forward} + \underbrace{\overleftarrow{L_{bl}^{s}} + \overleftarrow{L_{bl}^{e}}}_{backward}  + ||f(x^{f}) -  f(x^{b})||^{2},
\end{equation}
where $ \overrightarrow{L_{bl}^{s}} $ and $ \overrightarrow{L_{bl}^{e}} $ represent the $ L_{bl} $ between $ \overrightarrow{\mathbf{H}}^{s} $ and $ \mathbf{G}_{s}^{w} $, $ \overrightarrow{\mathbf{H}}^{e} $ and $ \mathbf{G}_{E}^{w} $ respectively in the forward pass. Mean-Squared Loss is also performed on two-passes intermediate features. 

\noindent
\textbf{Objective of PRB.} Taking the constructed proposal feature maps $\mathbf{F}^{p} $ as input, our PRB will generate two types of confidence maps $\mathbf{M}^{cr}$ and $\mathbf{M}^{cc}$ for all densely distributed proposals as \cite{LinBMN}. The training objective is defined as the regression loss $ L_{reg} $ and the binary classification loss $ L_{cls} $ respectively:
\begin{equation}
	L_{PRB} =  L_{reg} + L_{cls} ,
\end{equation}
where the smooth-$ L_{1} $ loss \cite{Girshick2015Fast} is adopted as  $ L_{reg} $, and the points  $ g^{c}_{i,j} $ with value large than 0.7 or lower than 0.3 are regarded as positives and negatives respectively. And we ensure the scale and number ratio between positives and negatives to be near 1:1 by the two-stage sampling scheme described above.

\subsection{Inference}
During inference stage, our BSN++ can generate the boundary map $ \mathbf{M}^{b} $ based on the bidirectional boundary probabilities ($ \mathbf{H}^{s}$ and $\mathbf{H}^{e}$) and confidence map ($ \mathbf{M}^{cc} $ and $ \mathbf{M}^{cr} $). We form the proposal map $ \mathbf{M}^{p} $ directly by fusing the $ \mathbf{M}^{b} $ and $ \mathbf{M}^{c} $ with dot multiplication. Then we can filter the points with high scores in the proposal map $ \mathbf{M}^{p} $ as candidate proposals used for post-processing.

\noindent
\textbf{Score Fusion.} As described above, the final scores of proposals in $ \mathbf{M}^{p} $ involve the local boundary information and global confidence scores. Take the proposal $\varphi$ = $[t_{s}, t_{e}]$ for example, the combination of final score $ p_{\varphi} $ can be shown as:
\begin{equation}
	\begin{aligned}
		p_{\varphi} & = \mathbf{M}^{b}_{t_{e}-t_{s},t_{s}} \cdot \sqrt{\mathbf{M}^{cc}_{t_{e}-t_{s},t_{s}} \cdot \mathbf{M}^{cr}_{t_{e}-t_{s},t_{s}}}, \\
	\end{aligned}
\end{equation}

\noindent
\textbf{Redundant Proposals Suppression.} BSN++ can generate the proposal candidates set as $ \mathbf{\Psi}_{p}= \{\varphi_{n}=(t_{s},t_{e}, p_{\varphi})\}_{n=1}^{N_{p}}$, where $ N_{p} $ is the number of proposals. Since the generated proposals may overlap with each other, we conduct Soft-NMS \cite{Bodla2017} algorithm to suppress the confidence scores of redundant proposals. Final, the proposals set is $ \mathbf{\Psi}_{p}'= \{\varphi_{n}^{'}=(t_{s},t_{e}, p_{\varphi}^{'})\}_{n=1}^{N_{p}}$, where $ p_{\varphi}^{'} $ is the decayed score of proposal $ \varphi_{n}^{'} $.


\section{Experiments}

\subsection{Datasets and Setup}

\textbf{Datasets.} \textbf{ActivityNet-1.3} \cite{Anet} is a large-scale video dataset for action recognition and temporal action detection tasks used in the ActivityNet Challenge from 2016 to 2020. It contains 19, 994 videos with 200 action classes temporally annotated, and the ratio of training, validation and testing sets is 1:1:2. \textbf{THUMOS-14} \cite{Jiang} contains 200 and 213 untrimmed videos with temporal annotations of 20 action classes in validation and testing sets respectively.

\noindent
\textbf{Implementation details.} For feature encoding, we adopt the two-stream network \cite{K.Simonyan}, where ResNet network \cite{K.He} and BN-Inception network \cite{S.Ioffe} are used as the spatial and temporal networks respectively. During feature extraction, the interval $ \sigma $ is set to 16 and 5 on ActivityNet-1.3 and THUMOS14 respectively. On ActivityNet-1.3, we rescale the feature sequence of input videos to $ l_{w}=100 $ by linear interpolation following \cite{BSN}, and the maximum duration $ D $ is also set to 100 to cover all action instances. While on THUMOS14, the length $ l_{w} $ of sliding windows is set to 128 while the maximum duration $ D $ is set to 64, which can cover almost 98\% action instances. On both datasets, we train our BSN++ from scratch using the Adam optimizer and the batch size is set to 16. And the initial learning rate is set to 0.001 for 7 epochs, then 0.0001 for another 3 epochs.


\subsection{Temporal Proposal Generation} 

\textbf{Evaluation metrics.} Following the conventions, 
Average Recall (AR) is calculated under different \textit{tIoU} thresholds which are set to [0.5:0.05:0.95] on ActivityNet-1.3, and [0.5:0.05:1.0] on THUMOS14. We measure the relation between AR and Average Number (AN) of proposals, denoted as AR@AN. And we also calculate the area (AUC) under the AR vs. AN curve as another evaluation metric on ActivityNet-1.3 dataset, where AN ranges from 0 to 100.


\noindent
\textbf{Comparison to the state-of-the-arts.} Table \ref{table_proposal_anet} illustrates the comparison results on ActivityNet-1.3. It can be observed that our BSN++ outperforms other state-of-the-art proposal generation methods with a big margin in terms of AR@AN and AUC on validation set of ActivityNet-1.3. For a direct comparison to BSN, our BSN++ improves AUC from 66.17\% to 68.26\% on validation set. Particularly, when the AN is 100, our method significantly improves AR from 74.16\% to 76.52\% by 2.36\%. And when the AN is 1, the AR which our BSN++ can obtain is 34.30\%.

\begin{figure}[t]
	\begin{center}
		\includegraphics[width=0.96\linewidth]{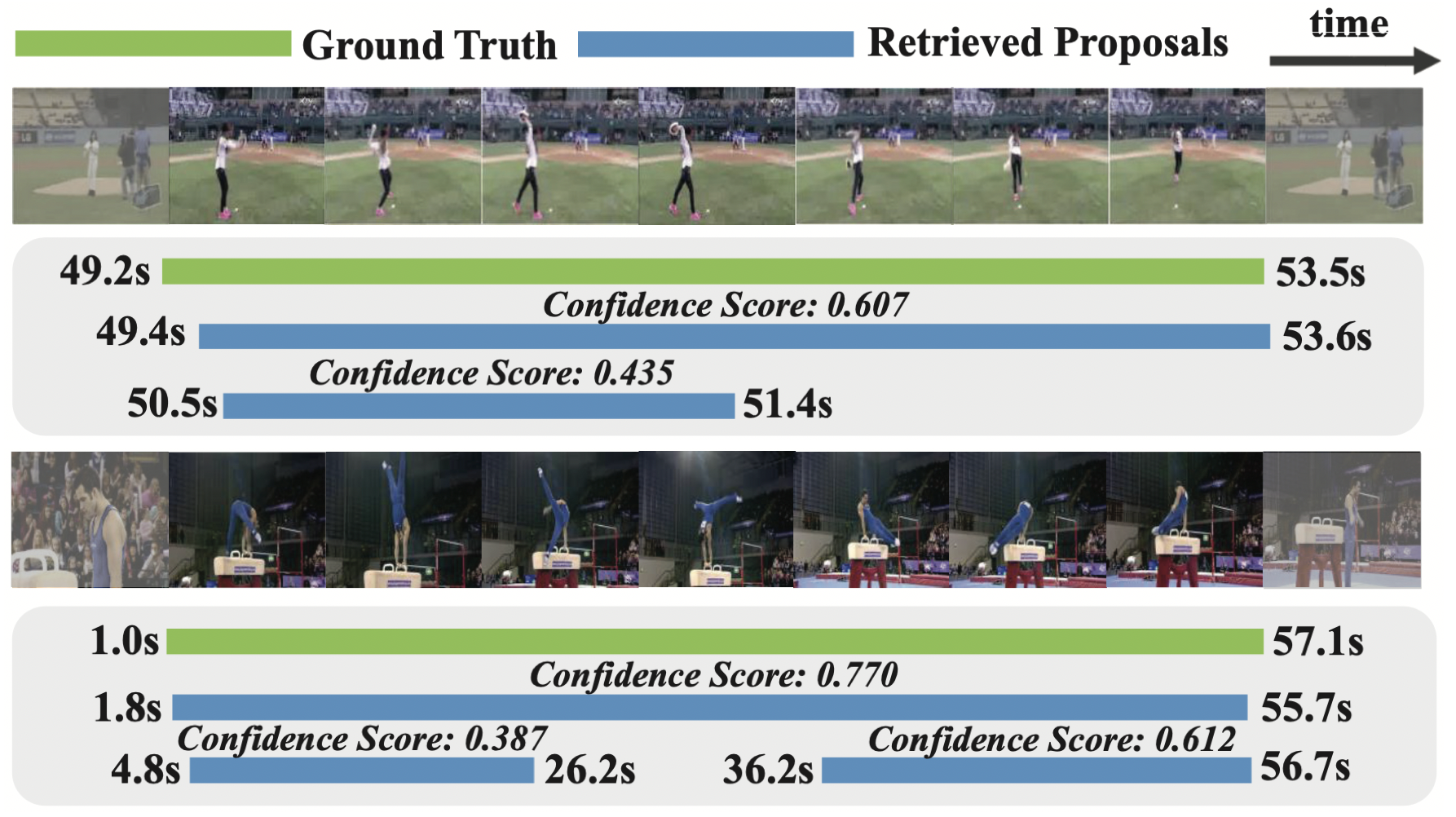}
	\end{center}
	\caption{Qualitative examples of proposals generated by BSN++ on THUMOS14 (top) and ActivityNet-1.3 (bottom).}
	\label{fig:qualitative_examples}
\end{figure}

Table \ref{table_proposal_thumos} illustrates the comparison results on THUMOS14 dataset. For fair comparisons, we use the features when compared with other methods, which mainly includes two-stream features and C3D features \cite{D.Tran}. Results shown in Table \ref{table_proposal_thumos} clearly demonstrate that: (1) the performance of our BSN++ obviously outperforms other state-of-the-methods in terms of AR@AN with AN varying from 50 to 1000, no matter what kind of features is served as input; (2) when post-processed with Soft-NMS, the higher AR can be obtained with fewer proposals. Qualitative examples on THUMOS14 and ActivityNet-1.3 are shown in Fig. \ref{fig:qualitative_examples}.
 
\setlength{\tabcolsep}{1pt}
\begin{table}[t]
	\caption{Performance comparisons with other state-of-the-art proposal generation methods on validation set of ActivityNet-1.3 in terms of AUC and AR@AN.}
	\begin{tabular}{m{2.1cm}cccccc}
		\toprule
		Method  & SSAD-prop &CTAP &BSN &MGG & BMN & \textbf{BSN++} \tabularnewline
		\noalign{\smallskip}
		\hline  
		\noalign{\smallskip}
		AR@1 (val) & -  & - & 32.17 & - & - & \textbf{34.30}  \tabularnewline
		AR@100 (val) & 73.01 & 73.17 & 74.16 & 74.54 & 75.01 & \textbf{76.52} \tabularnewline
		AUC (val) & 64.40 & 65.72 & 66.17 & 66.43 & 67.10 &  \textbf{68.26} \tabularnewline
 
		\bottomrule
	\end{tabular}
	\label{table_proposal_anet}
\end{table}

\setlength{\tabcolsep}{4pt}
\begin{table}[tbp]
	\centering
	\caption{Comparisons with other state-of-the-art proposal generation methods SCNN-prop\cite{SCNN}, SST\cite{sst_buch_cvpr17}, TURN\cite{gao2017turn}, MGG\cite{LiuMulti}, BSN\cite{BSN}, BMN\cite{LinBMN}  on THUMOS14 in terms of AR@AN, where SNMS stands for Soft-NMS.}
	\small
	\begin{tabular}{m{1.3cm}m{2cm}m{0.6cm}m{0.6cm}m{0.6cm}m{0.6cm}m{0.8cm}}
		\toprule
		Feature & Method  & @50  & @100 & @200 & @500 & @1000    \\
		\noalign{\smallskip}
		\hline
		\noalign{\smallskip}
		C3D & TURN & 19.63 & 27.96 & 38.34 & 53.52 & 60.75  \\
		C3D & MGG  & 29.11 & 36.31 & 44.32 & 54.95 & 60.98 \\
		C3D & BSN(SNMS) & 29.58 & 37.38 & 45.55 & 54.67 & 59.48 \\ 
		C3D & BMN(SNMS) & 32.73 & 40.68 & 47.86 & 56.42 & 60.44 \\
		\noalign{\smallskip}
		\hline
		\noalign{\smallskip}
		C3D & \textbf{BSN++}(SNMS) & \textbf{34.88}  & \textbf{43.72} & \textbf{50.12}  & \textbf{58.88} & \textbf{61.39}\\
		\noalign{\smallskip}
		\hline
		\hline
		\noalign{\smallskip}
		2-Stream & CTAP & 32.49 & 42.61 & 51.97 & - & - \\
		2-Stream & MGG & 39.93 & 47.75 & 54.65 & 61.36 & 64.06 \\
		2-Stream & BSN(SNMS)& 37.46 & 46.06 & 53.21 & 60.64 & 64.52 \\
		2-Stream & BMN(SNMS)& 39.36 & 47.72 & 54.70 & 62.07 & 65.49 \\
		\noalign{\smallskip}
		\hline
		\noalign{\smallskip}
		2-Stream & \textbf{Ours}(SNMS) & \textbf{42.44} & \textbf{49.84} & \textbf{57.61} & \textbf{65.17} & \textbf{66.83}   \\
		\bottomrule
	\end{tabular}
	\label{table_proposal_thumos}
\end{table}

\setlength{\tabcolsep}{6pt}
\begin{table}[t]
	\begin{center}
		\caption{Ablation experiments in the validation set of ActivityNet-1.3. Complementary boundary generator is abbreviated as CBG and BBM denotes bi-directional matching. PRB is the proposal relation block and SBS is the scale-balanced sampling. PAM and CAM indicate the two self-attention modules. Inference speed here is the seconds (s) cost $T_{cost} $ for processing a 3-minute videos using a Nvidia 1080-Ti card. e2e denotes the joint training manner. }
		\begin{tabular}{p{1cm}ccccccc}
			\toprule
			Model & Module & e2e & AUC & $T_{cost} $ \tabularnewline
			\noalign{\smallskip}
			\hline  
			\noalign{\smallskip}
			BSN & TEM & - & 64.80 & 0.036 \tabularnewline
			BSN & TEM+PEM & $ \times $  & 66.17 & 0.629 \tabularnewline
			BMN & TEM & - & 65.17  & 0.035  \tabularnewline
			BMN & TEM+PEM & $ \checkmark $ & 67.10  & 0.052 \tabularnewline
			\noalign{\smallskip}
			\hline  
			\noalign{\smallskip}
			BSN++ & CBG(w/o BBM) & -  & 66.02  & 0.019  \tabularnewline
			BSN++ & CBG(w/ BBM) & -  & 66.43  & 0.025 \tabularnewline
			BSN++ & CBG+PRB (w/o SBS) & $ \times $  & 67.34  & 0.054 \tabularnewline
			BSN++ & CBG+PRB (w/o SBS) & $  \checkmark $  & 67.77  & 0.039 \tabularnewline
			BSN++ & CBG+PRB (w/o PAM) & $  \checkmark $  & 67.99   & 0.034 \tabularnewline
			BSN++ & CBG+PRB (w/o CAM) & $  \checkmark $  & 68.01   & 0.035 \tabularnewline
			BSN++ & CBG+PRB  & $  \checkmark $ & \textbf{68.26}  & \textbf{0.039} \tabularnewline
			
			\bottomrule 
		\end{tabular}
		\label{table_ablation_anet}
	\end{center}
\end{table}

  \setlength{\tabcolsep}{8pt}
  \begin{table}[t]
  	\centering
  	\caption{Generalizability evaluation on ActivityNet-1.3.}
  	\small
  	\begin{tabular}{p{1.4cm}p{1.1cm}<{\centering}p{1.1cm}<{\centering}p{1.1cm}<{\centering}p{1.2cm}<{\centering}}
  		\toprule
  		BMN/BSN++ & \multicolumn{2}{c}{Seen(validation)} & \multicolumn{2}{c}{Unseen(validation)}   \\
  		\hline
  		Train Data  & AR@100 &  AUC  & AR@100  & AUC  \\
	 	\hline
  		Seen+Unseen & 72.96/74.56 &  65.02/66.34 &  72.68/74.32 & 65.06/66.37\\
  		Seen & 72.47/74.03 & 64.37/65.87  & 72.46/\textbf{73.82} & 64.47/\textbf{65.89} \\
  		\bottomrule
  	\end{tabular}
  	\label{generalizability}
  \end{table}

 \setlength{\tabcolsep}{10pt}
\begin{table}[t]
	\centering
	\caption{Detection results compared with \cite{SSN,SSAD,BSN,LinBMN,PGCN,GTAD} on validation set of ActivityNet-1.3, where our proposals are combined with video-level classification results generated by \cite{xiong2016cuhk}.}
	\small
	\begin{tabular}{p{1.5cm}p{0.62cm}<{\centering}p{0.62cm}<{\centering}p{0.62cm}<{\centering}p{0.9cm}<{\centering}}
		\toprule
		\multicolumn{5}{c}{ {\bf ActivityNet-1.3}, mAP@$tIoU$}  \\
		\hline
		& \multicolumn{4}{c}{validation} \\
		\hline
		Method  & 0.5  &  0.75  & 0.95  & Average  \\
		\hline
		SSN    & 39.12 & 23.48  & 5.49  & 23.98\\
		SSAD & 44.39  & 29.65  & 7.09  & 29.17 \\
		BSN & 46.45 & 29.96 & 8.02 & 30.03\\
		P-GCN & 48.26 & 33.16 & 3.27 & 31.11 \\ 
		BMN & 50.07 & 34.78 & 8.29 & 33.85\\
		G-TAD & 50.36 & 34.60 & \textbf{9.02} & 34.09   \\
		\hline
		\textbf{Ours} & \textbf{51.27}  & \textbf{35.70}  & 8.33  & \textbf{34.88} \\
		\bottomrule
	\end{tabular}
	\label{table_detection_anet}
\end{table}

\setlength{\tabcolsep}{5pt}
\begin{table}[t]
	\centering
	\caption{Detection results compared with \cite{gao2017turn,BSN,LiuMulti,LinBMN} on testing set of THUMOS14, where video-level classifier \cite{L.Wang} is combined with proposals generated by BSN++.}
	\small
	\begin{tabular}{p{2cm}|p{1.1cm}<{\centering}|p{0.5cm}<{\centering}p{0.5cm}<{\centering}p{0.5cm}<{\centering}p{0.5cm}<{\centering}p{0.5cm}<{\centering}}
		\toprule
		\multicolumn{7}{c}{ {\bf THUMOS14} (testing), mAP@$tIoU$}  \\
		\hline
		\noalign{\smallskip}
		Method & Classifier & 0.7 & 0.6 & 0.5 & 0.4 & 0.3  \\
		\noalign{\smallskip}
		\hline
		\noalign{\smallskip}
		TURN & UNet	& 6.3 & 14.1 & 24.5 & 35.3 &  46.3\\
		BSN & UNet & 20.0 & 28.4 & 36.9 & 45.0 & 53.5\\
		MGG & UNet & 21.3 & 29.5 & 37.4 & 46.8 & 53.9\\
		BMN & UNet & 20.5 & 29.7 & 38.8 & 47.4 & 56.0\\
		\noalign{\smallskip}
		\hline
		\noalign{\smallskip}
		\textbf{Ours} & UNet & \textbf{22.8} & \textbf{31.9} & \textbf{41.3} & \textbf{49.5} & \textbf{59.9}\\
		\bottomrule
	\end{tabular}
	\label{table_detection_thumos}
	\normalsize
	
\end{table}

\subsection{Ablation Experiments}
In this section, we comprehensively evaluate our proposed BSN++ on the validation set of ActivityNet-1.3.

\noindent
\textbf{Effectiveness and efficiency of modules in BSN++.} We perform the ablation studies with different architecture settings to verify the effectiveness and efficiency of each module proposed in BSN++. The evaluation results shown in Table \ref{table_ablation_anet} demonstrate that: (1) the Encoder-Decoder architecture can effectively learn ``\textit{local and global}" contexts for accurate boundary prediction compared to the previous works which only explore the local details; (2) the bidirectional matching mechanism further validates the importance of future context in assisting the boundary judgement; (3) unlike the previous works which treat the proposals separately, the proposal relation block can provide more comprehensive features for accurate and discriminative proposals scoring; (4) besides, with scale-balanced sampling, the model can obtain equivalent balancing; (5) final, integrating all the separated modules into an end-to-end network, we can obtain the competing performance improvement; (6) BSN++ achieves the great overall efficiency than previous methods.

\noindent
\textbf{Ablation comparison with BSN.} We conduct a direct comparison to BSN\cite{BSN} to confirm the effectiveness and superiority of our BSN++. As shown in Table \ref{table_ablation_anet}, the TEM of BSN which only considers the local details for boundary probabilities sequence generation is inferior with limited receptive fields. Meanwhile, without the full usage of temporal context, it is also not robust in complicated scenarios. Besides, BSN fails to model the proposal relations for confidence regression, as well as neglect the imbalance data distribution against proposal duration. However, our BSN++ handles these issues accordingly and effectively.

\noindent
\textbf{Generalizability of proposals.} Another key property of the proposal generation method is the generalizability. To evaluate this property, two un-overlapped action subsets: ``Sports, Exercise, and Recreation" and ``Socializing, Relaxing, and Leisure" of ActivityNet-1.3 are chosen as \textit{seen} and \textit{unseen} subsets separately. There are 87 and 38 action categories, 4455 and 1903 training videos, 2198 and 896 validation videos on \textit{seen} and \textit{unseen} subsets separately. We adopt C3D network pre-trained on Sports-1M dataset for feature extraction. Then we train BSN++ with \textit{seen} and \textit{seen+unseen} training videos separately, and evaluate both models on \textit{seen} and \textit{unseen} validation videos separately. Results in Table \ref{generalizability} reveal that there is only slight performance drop on unseen categories, suggesting that BSN++ achieves great generalizability to generate high quality proposals for unseen actions.

\subsection{Action Detection with Our Proposals}
\noindent
\textbf{Evaluation metrics.} For temporal action detection task, mean Average Precision (mAP) is a conventional evaluation metric, where Average Precision (AP) is calculated for each action category respectively. On ActivityNet-1.3, the mAP with \textit{tIoU} thresholds set $ \{0.5, 0.75, 0.95\} $ and the average mAP with \textit{tIoU} thresholds [0.5:0.05:0.95] are reported. On THUMOS14, mAP with \textit{tIoU} thresholds set $\{0.3, 0.4, 0.5, 0.6, 0.7\}$ is used.

\noindent
\textbf{Comparison to the state-of-the-arts.}
To further examine the quality of proposals generated by BSN++, following BSN\cite{BSN}, we feed them to the state-of-the-art action classifiers to obtain the categories for action detection in a \textit{``detection by classification"} framework. On ActivityNet-1.3, we use the top-1 video-level classification results generated by \cite{xiong2016cuhk} for all the generated proposals. And on THUMOS14, we use the top-2 video-level classification results generated by UntrimmedNet \cite{L.Wang}. Comparison results are illustrated in Table \ref{table_detection_anet} and Table \ref{table_detection_thumos} respectively. We can observe that with the same classifiers, the detection performance of our method can be boosted greatly, which can further demonstrate the effectiveness and superiority of our method.

\section{Conclusion}

We propose BSN++ for temporal action proposal generation. The complementary boundary generator takes the advantage of U-shaped architecture and bi-directional boundary matching mechanism to learn rich contexts for boundary prediction. To model the proposal-proposal relations for confidence evaluation, we devise the proposal relation block which employs two self-attention modules to perform global and inter-dependencies modeling. Meanwhile, we are the first to consider the imbalanced data distribution of proposal durations. Both the boundary map and confidence map can be generated simultaneously in a unified network. Extensive experiments conducted on ActivityNet-1.3 and THUMOS14 datasets demonstrate the effectiveness of our method in both temporal action proposal and detection performance.

%
%
\bibliographystyle{splncs04}
\bibliography{egbib}
\end{document}